\documentclass[conference,a4paper]{IEEEtran}
\IEEEoverridecommandlockouts

\usepackage[hidelinks]{hyperref}
\usepackage[cmex10]{amsmath}
\usepackage{amssymb,amsfonts}
\usepackage{dblfloatfix}

\usepackage[ruled,vlined]{algorithm2e}
\usepackage{graphicx}

\usepackage{booktabs}
\usepackage{siunitx}
\usepackage[numbers,compress]{natbib}
\usepackage{texnames}
\usepackage{bm,bbm}
\usepackage{orcidlink}

\begin{document}
\title{{Semantic-CD: Remote Sensing Image Semantic Change Detection towards Open-vocabulary Setting}
\thanks{$^\star$Corresponding Author}}

\author{Yongshuo~Zhu,~Lu~Li$^{\star}$,~Keyan~Chen,~Chenyang~Liu,~Fugen~Zhou,~Zhenwei~Shi\\
Beihang University}

\maketitle

\setlength{\abovedisplayskip}{6pt}
\setlength{\belowdisplayskip}{2pt}

\begin{abstract}

Remote sensing image semantic change detection is a method used to analyze remote sensing images, aiming to identify areas of change as well as categorize these changes within images of the same location taken at different times.
Traditional change detection methods often face challenges in generalizing across semantic categories in practical scenarios. To address this issue, we introduce a novel approach called Semantic-CD, specifically designed for semantic change detection in remote sensing images. This method incorporates the open vocabulary semantics from the vision-language foundation model, CLIP. By utilizing CLIP's extensive vocabulary knowledge, our model enhances its ability to generalize across categories and improves segmentation through fully decoupled multi-task learning, which includes both binary change detection and semantic change detection tasks. Semantic-CD consists of four main components: a bi-temporal CLIP visual encoder for extracting features from bi-temporal images, an open semantic prompter for creating semantic cost volume maps with open vocabulary, a binary change detection decoder for generating binary change detection masks, and a semantic change detection decoder for producing semantic labels. Experimental results on the SECOND dataset demonstrate that Semantic-CD achieves more accurate masks and reduces semantic classification errors, illustrating its effectiveness in applying semantic priors from vision-language foundation models to SCD tasks.
\end{abstract}

\begin{IEEEkeywords}
 Remote sensing image, semantic change detection, foundation model, open-vocabulary
\end{IEEEkeywords}

\IEEEpeerreviewmaketitle

\section{Introduction}

Semantic Change Detection (SCD) in remote sensing imagery aims to identify and categorize changes between images captured at different times of the same location. This technique is crucial in fields such as environmental monitoring, urban development, and disaster management. However, most existing methods focus solely on binary change detection (BCD), which indicates whether a change has occurred but does not specify the nature of the change \cite{li2022remote,lei2021difference,rahman2018siamese,hou2019w,daudt2018fully,chen2020dasnet,jiang2020pga,chen2020spatial,chen2021remote}. Typically, these methods concentrate on single change types, like buildings, thereby limiting their practical applicability.

To address these limitations, semantic change detection approaches in remote sensing offer a more general understanding by identifying specific change types beyond BCD results \cite{ding2024joint,chen2024changemamba,niu2023smnet,peng2021scdnet,daudt2019multitask,ding2022bi,yang2021asymmetric,zheng2022changemask,tian2023temporal}. These methods achieve accurate change semantics through strategies such as spatiotemporal consistency \cite{ding2024joint,chen2024changemamba}, multi-scale feature fusion \cite{niu2023smnet,peng2021scdnet}, and multi-task decoupling and integration \cite{daudt2019multitask,ding2022bi,yang2021asymmetric,zheng2022changemask}. Despite some progress with current remote sensing image SCD techniques, they often operate within closed set categories, restricting their ability to handle diverse semantic groups. Introducing open vocabulary concepts presents a new framework for tackling the generalization challenge. However, the field of SCD lacks the datasets with the extensive category coverage necessary for open vocabulary research. Therefore, exploring open vocabulary contexts in change detection remains significant. This study explores a degraded open-vocabulary SCD framework and conducts experimental evaluation, showing the model's potential in open vocabulary application. We propose Semantic-CD, a method that decouples change detection into BCD and SCD tasks. The BCD task utilizes the general visual semantic understanding capabilities of the vision-language foundation model CLIP \cite{radford2021learning} for binary segmentation, while the SCD task leverages CLIP’s open vocabulary capabilities to interpret semantic changes. The principal contributions of this paper include:

i) Leveraging the generalization capabilities of the vision-language foundational model CLIP, this paper introduces a semantic change detection framework for remote sensing images with open vocabulary settings, termed Semantic-CD.

ii) We fully decompose change detection into two tasks: BCD and SCD. By integrating CLIP's visual priors, we directly generate a BCD mask. Furthermore, we propose an instance-level semantic-guided class prompter to enable open vocabulary encoding, with the CLIP text encoder integrating semantic knowledge with bi-temporal visual features, constructing a semantic cost map that facilitates semantic change detection.

iii) Experimental results on the SECOND dataset demonstrate that Semantic-CD generates more accurate semantic change detection masks, achieving state-of-the-art performance. They demonstrate the efficacy of vision-language foundational model priors in SCD tasks.

\begin{figure*}[htbp]
    \centering
    \includegraphics[width=0.9\linewidth]{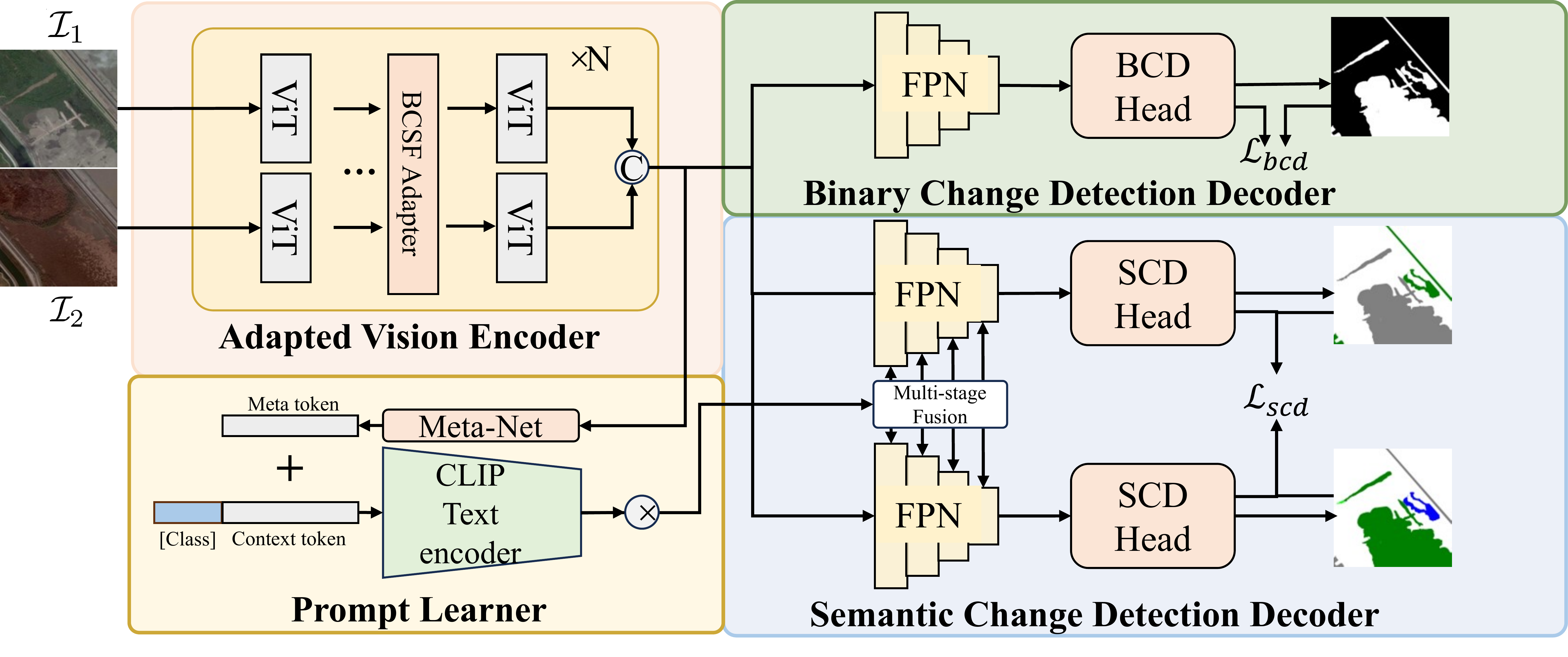}
    \caption{The architecture of the Semantic-CD consists of four main components:  an adapted CLIP vision encoder for extracting robust features from bi-temporal images; a Prompt Learner tasked with generating visual embeddings that encapsulate textual semantics; a BCD decoder responsible for producing binary detection masks; and a SCD decoder which assigns precise semantic labels to these binary masks.}
    \label{fig:model}
\end{figure*}

\section{Methodology}

\subsection{Overall Architecture}

The Semantic-CD consists of four key components: the Bi-temporal CLIP visual encoder, the open semantic prompter, the binary Change Detection (BCD) decoder, and the Semantic Change Detection (SCD) decoder, as depicted in Fig. \ref{fig:model}. The entire process is mathematically described by the following,

\begin{align}
\begin{split}
    (F_1, F_2) &= \Phi_{\text{CLIP-enc}} (\mathcal{I}_1, \mathcal{I}_2)
    \\
    (F_{\text{cv}}^1, F_{\text{cv}}^2) &= \Phi_{\text{prompt-learner}} (\mathbb{C}, F_1, F_2) \\
    M_{\text{BCD}} &= \Phi_{\text{BCD-decoder}} (F_1, F_2)
    \\
    (M_{\text{SCD}}^1, M_{\text{SCD}}^2) &= \Phi_{\text{SCD-decoder}} (F_1, F_2, F_{\text{cv}}^1, F_{\text{cv}}^2)
\end{split}
\end{align}
where the pre-temporal ($\mathcal{I}_1 \in \mathbb{R}^{H \times W \times 3}$) and post-temporal ($\mathcal{I}_2 \in \mathbb{R}^{H \times W \times 3}$) images are fed into the bi-temporal CLIP visual encoder, denoted as $\Phi_{\text{CLIP-enc}}$. This operation produces feature maps $F_1 \in \mathbb{R}^{N \times C}$ and $F_2 \in \mathbb{R}^{N \times C}$. For binary change detection, these features are fed into the BCD decoder, $\Phi_{\text{BCD-decoder}}$, resulting in a binary change detection mask $M_{\text{BCD}} \in \mathbb{R}^{H \times W}$. 
For semantic change detection, the features $F_1$, $F_2$, and a class set $\mathbb{C}$ are initially processed by the prompt learner, $\Phi_{\text{prompt-learner}}$. This step yields pixel-level cost volumes $F_\text{cv}^1 \in \mathbb{R}^{H' \times W' \times C}$ and $F_\text{cv}^2 \in \mathbb{R}^{H' \times W' \times C}$, where $C$ denotes the number of classes. The SCD decoder, $\Phi_{\text{SCD-decoder}}$, subsequently combines these cost volumes with $F_1$ and $F_2$ to generate semantic change detection masks, $M_\text{SCD}^1 \in \mathbb{R}^{H \times W \times C}$ and $M_\text{SCD}^2 \in \mathbb{R}^{H \times W \times C}$.

\subsection{Bi-temporal CLIP visual encoder,}

Change detection seeks to identify effective changes in bi-temporal images while excluding invalid changes arising from external factors such as varying lighting conditions, atmospheric effects, and calibration inconsistencies. This paper harnesses the general capabilities of the CLIP foundation model for semantic extraction. Despite its utility, the CLIP visual encoder struggles with discerning valid changes in bi-temporal imagery. Zhu et al. \cite{zhu2024semantic} proposed the Bi-temporal Change Semantic Filter (BCSF) within the SAM framework to enhance the foundation model's ability to express change-related semantic features across spatial and channel dimensions, yielding excellent results. This paper incorporates BCSF into the CLIP image encoder, functioning as a change semantic adapter to improve its ability to capture and comprehend valid changes in bi-temporal images. 
The BCSF is specifically applied to each of the four global attention layers within the visual backbone. During model training, all other backbone parameters remain fixed, with only the parameters associated with the BCSF layers being fine-tuned.

\subsection{Open Semantic Prompter}

The CLIP text encoder, having been trained on millions of image-text data pairs, CLIP text encoder possesses generalized semantic knowledge. In this paper, we develop an open-vocabulary prompter based on instance-level change semantics. This prompt comprises both category vocabulary embeddings and context embeddings. The context embeddings are constructed from task-specific learnable tokens and meta tokens, which are informed by instance-level change semantics, as illustrated in Fig. \ref{fig:model}. The open-vocabulary capabilities are demonstrated by the CLIP text encoder's ability to encode text from any category. The encoding process using the open-vocabulary prompt is detailed as follows:
\begin{align}
    \begin{split}
        p{_i}^{\text{m}} &=  \Phi_{\text{meta}} (F_i), i \in\{1,2\} 
        \\
        p{_i}^{\text{c}} &= p{_i}^{\text{m}} + p^{\text{t}}
        \\
        t_{i,j} &= \Phi_{\text{cat}}(p{_i}^{\text{c}} + p{_j}^{\text{text}}), j \in\mathbb{C} \\
        t_{i,j} &=\Phi_{\text{CLIP-text}}(t_{i,j}) 
        \\
        F_\text{cv}^i&= \Phi_{\text{sim}}(F_i, t_{i,j})
    \end{split}
\end{align}
where the instance-level change semantic meta tokens, denoted as \( p_{i}^{\text{m}} \), are derived from \( F_i \) using the meta network \(\Phi_{\text{meta}}\). The task-specific learnable tokens, \( p^{\text{t}} \), comprise a set of globally shared vectors. The context semantic embeddings, \( p_{i}^{\text{c}} \), are obtained by summing the meta tokens and the learnable tokens. The context embeddings are then appended to the category embeddings \( p_{j}^{\text{text}} \) and input into the CLIP text encoder, \(\Phi_{\text{CLIP-text}}\), to produce semantic embeddings \( t_{i,j} \) for different temporal phases (\( i \in \{1, 2\} \)) and category words (\( j \in \mathbb{C} \)). Finally, the cosine similarity with the original bi-temporal feature map is calculated to generate the cost volume map \( F_{\text{cv}}^i \in \mathbb{R}^{H' \times W' \times C} \) for various category semantics.

\subsection{BCD Decoder}

The binary change detection decoder employs a compact multi-scale change detection decode head, as illustrated in \cite{zhu2024semantic}, to decode bi-temporal visual features and produce a binary change mask, as described by the following equations:
\begin{align}
    \begin{split}
        \{ F_i' \} &= \Phi_{\text{sampling}} ( \Phi_{\text{cat}} ( F_1, F_2) ) \\
        F_i' &= \Phi_{\text{resize}} (\Phi_{\text{conv}} (F_i') ) \\
        M_\text{BCD} &= \Phi_{\text{proj}} (\Phi_{\text{cat}} (\{ F_i' \} ) ) \\
    \end{split}
\end{align}
where $F_1$ and $F_2$ denote the semantic feature maps extracted by the visual encoder for pre-change and post-change temporal states, respectively. $\Phi_{\text{sampling}}$ represents the multi-scale construction process based on SimpleFPN \cite{li2022exploring}. The resulting multi-scale feature maps, $\{ F_i' \in \mathbb{R}^{\frac{H}{2^{i+1}} \times \frac{W}{2^{i+1}} \times 512} \}, i \in \{1, 2, 3, 4 \}$, are spatially refined through convolutional operations. The function $\Phi_{\text{resize}}$ employs bilinear interpolation to unify the spatial resolution, while $\Phi_{\text{proj}}$ utilizes a $1 \times 1$ convolutional layer to map the feature maps' channel dimension to the number of classes.

\subsection{SCD Decoder}

\begin{figure}[!htbp]
    \centering
    \includegraphics[width=0.7\linewidth]{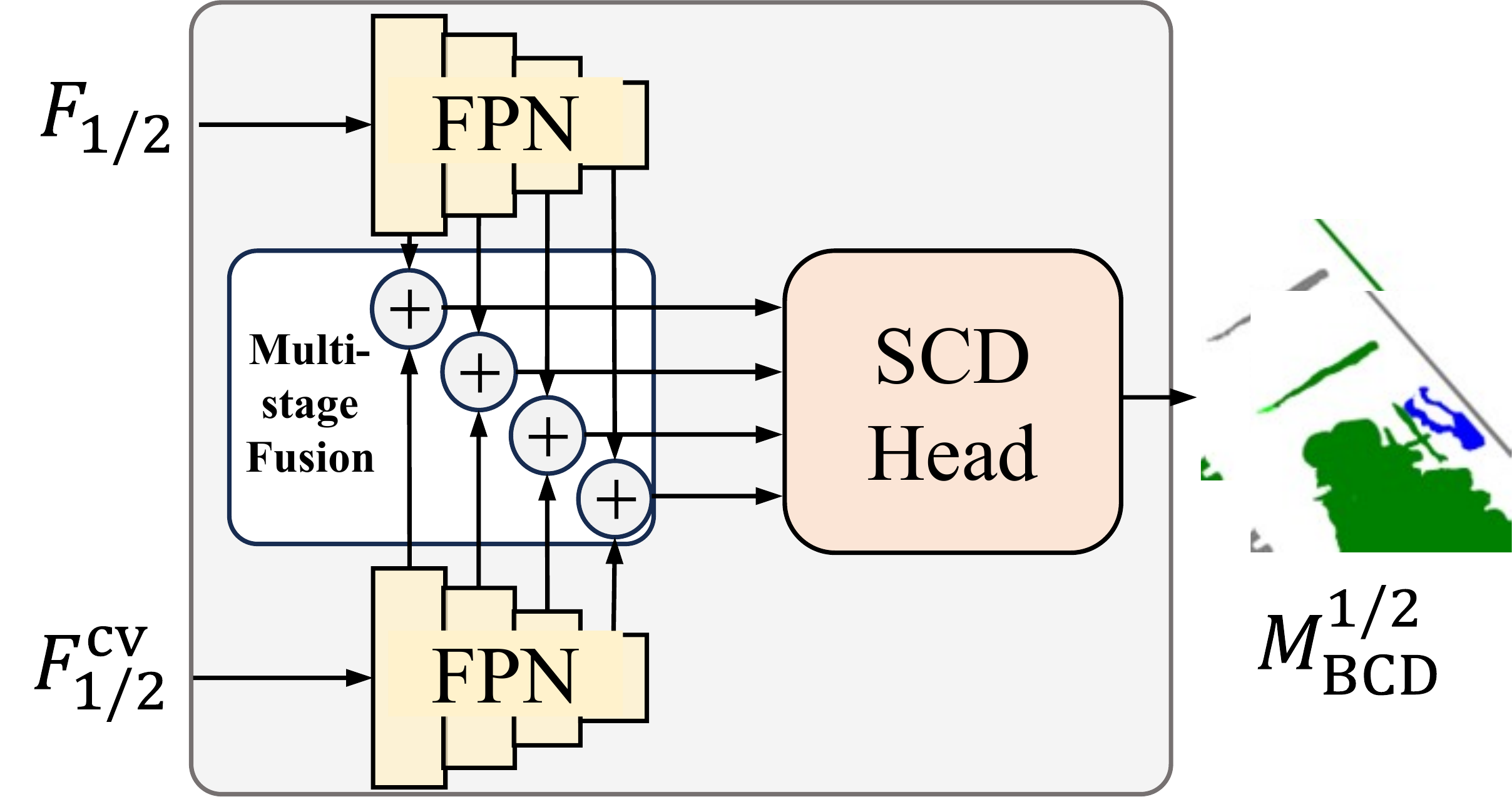}
    \caption{The structure of the Semantic Change Detection Decoder.}
    \label{fig:SCD}
\end{figure}

The SCD decoder comprises two identical decoder networks for processing visual features and a cost map-guided multi-stage fusion network, as depicted in Fig. \ref{fig:SCD}. Similar to the BCD decoder, bi-temporal features are independently fed into a SimpleFPN with shared parameters. Simultaneously, the cost map corresponding to different temporal phases is processed by another SimpleFPN to generate multi-level responses, as expressed in the following equations,
\begin{align}
    \begin{split}
        \{ F_i \} &= \Phi_{\text{sampling}} ( F_{i} ) \\
        \{ F_i^\text{cv} \} &= \Phi_{\text{sampling}} ( F_i^{cv} ) \\
    \end{split}
\end{align}
where $\{F_i\}$ and $\{F^i_{\text{cv}}\}$ denote the multi-scale bi-temporal semantic feature maps and cost maps produced by various SimpleFPN modules, respectively. Subsequently, these maps are hierarchically fused and unified into a fixed scale followed by a decoding mapping layer to generate the final semantic change mask, as follows,
\begin{align}
    \begin{split}
        F_i'  &= F^i_{\text{cv}} + F_i \\
        F_i' &= \Phi_{\text{resize}} (\Phi_{\text{conv}} (F_i') ) \\
        M_\text{BCD}^{1/2} &= \Phi_{\text{proj}} (\Phi_{\text{cat}} (\{ F_i' \} ) ) \\
    \end{split}
\end{align}
where $\{F_i'\}$, where $i \in \{1, 2, 3, 4 \}$, represents the fused multi-scale feature maps. The generation of the SCD mask from these feature maps follows a procedure similar to that used in the BCD decoder.

\subsection{Training Strategy} \label{sec:training_strategy}

The paper employs cross-entropy loss for supervised training, as detailed below,
\begin{align}
\begin{split}
    \mathcal{L}_{\text{bcd}} &= -\frac{1}{N}\sum_{i=1}^{N} \left( y_i^{\text{bcd}} \log (\hat{y}_i^{\text{bcd}}) + (1-y_i^{\text{bcd}}) \log (1 - \hat{y}_i^{\text{bcd}}) \right), \\
\mathcal{L}_{\text{scd}} &= -\frac{1}{N}\sum_{i=1}^{N}\sum_{c=1}^{\mathbb{C}-1} y_{i,c}^{\text{scd-1/2}} \log (\hat{y}_{i,c}^{\text{scd-1/2}})
\end{split}
\end{align}
where $^\text{bcd}$ refers to the binary change detection mask, while $^\text{scd-1/2}$ denote the pre-/post-temporal semantic change detection mask. The symbols $\hat{y}_i$ and $y_i$ denote the predicted and ground truth values of a single pixel, respectively. The variable $N$ signifies the total number of pixels. Our objective is to maximize the differentiation between BCD and SCD tasks. Consequently, we adopt a fully decoupled two-stage training paradigm: i) During the training for the BCD branch, the prompter and the SCD decoder are kept fixed; (ii) During the training for the SCD branch, the visual encoder and the BCD decoder are maintained in a frozen state.

\section{Experimental Results and Analyses}
\subsection{Experimental Setup}

\subsubsection{Dataset and Evaluation Metrics}

The experiments utilized the SECOND dataset \cite{yang2020semantic} for both training and evaluation. This dataset covers six primary land cover categories, capturing a range of typical natural and human-made changes. Consistent with related studies \cite{chen2024changemamba}, we employed four metrics to evaluatd the methods: Overall Accuracy (OA), F1 Score, Mean Intersection over Union (mIoU), and Separated Kappa coefficient (SeK).

\subsubsection{Implementation Details}

We conducted experiments using the RemoteCLIP pre-trained vision-language model, CLIP-ViT-L-14 \cite{liu2024remoteclip}. For the BCD training, we employed a batch size of 1 and trained for 300 epochs. In contrast, for the SCD, training was conducted for 5 epochs with a context token length of 256. To evaluate the effect of introducing an open vocabulary semantic prompter, we developed a baseline model without it, which significantly hindered the convergence of SCD training. As a result, we extended the SCD training to 100 epochs while maintaining a batch size of 1. This experimentation was implemented using PyTorch and evaluated on the NVIDIA A100 hardware platform.

\subsection{Experimental Results}

\begin{figure}[!htbp]
    \centering
    \includegraphics[width=1\linewidth]{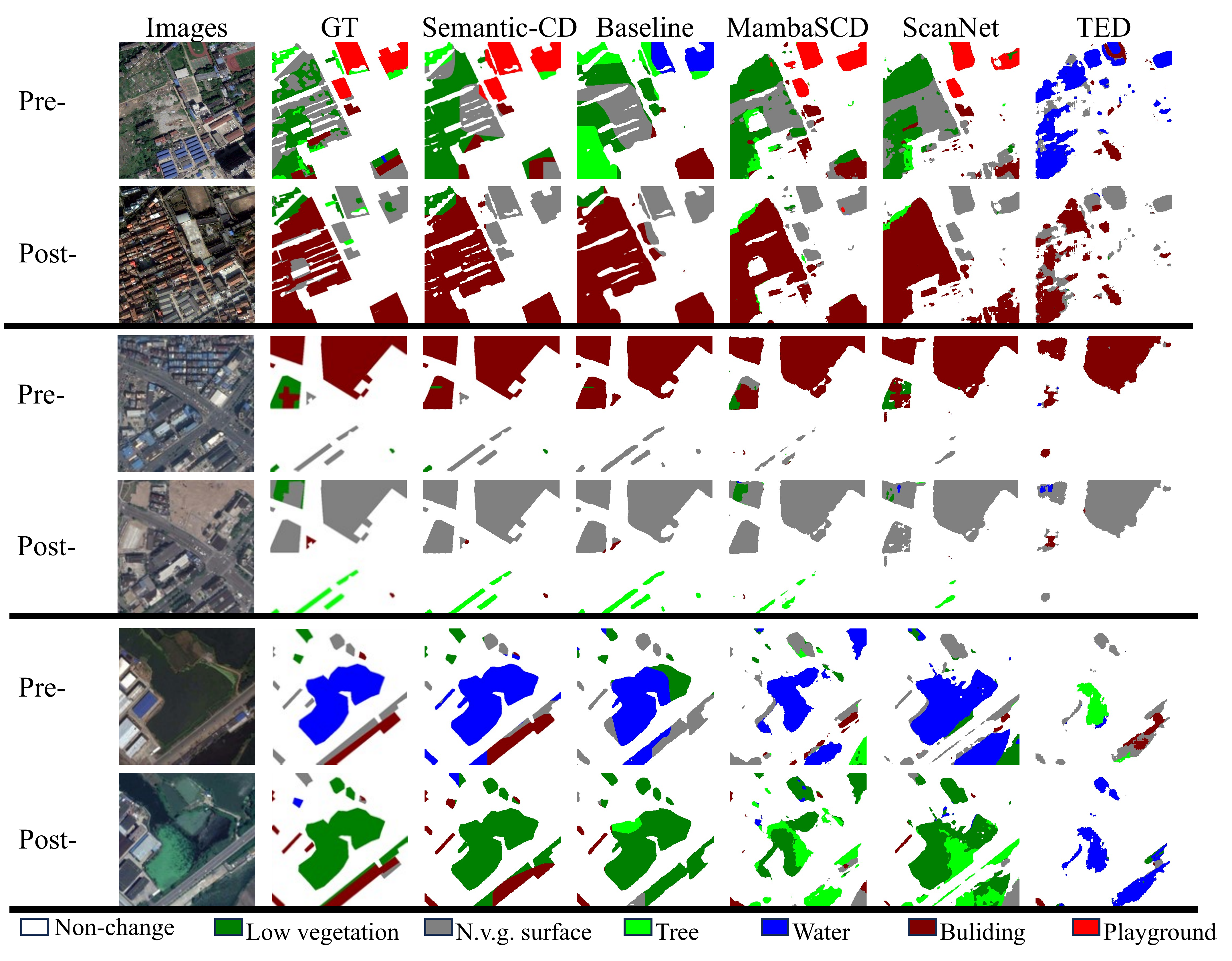}
    \caption{Qualitative comparison results with other methods on the SECOND dataset.
    }
    \label{fig:comp}
\end{figure}

The experimental results, both quantitative and qualitative, are presented in Fig. \ref{fig:comp} and Table. \ref{tab:comp}. In the quantitative analysis, we evaluate Semantic-CD against other state-of-the-art methodologies, as detailed in Table \ref{tab:comp}. The methods compared include several CNN-based approaches, such as HRSCD \cite{daudt2019multitask}, ChangeMask \cite{zheng2022changemask}, SSCD \cite{ding2022bi}, BiSRNet \cite{ding2022bi}, and TED \cite{ding2024joint}. Transformer-based methods like SMNet \cite{niu2023smnet} and ScanNet \cite{ding2024joint}, as well as Vmamba-based methods like MamabaSCD \cite{chen2024changemamba}, are also included. The results reveal that Semantic-CD consistently outperforms both the baseline and other competing methods across all evaluated metrics, underscoring its superior performance and the efficacy of semantic information extraction. Additionally, Fig. \ref{fig:comp} showcases qualitative visual segmentation results featuring methods such as TED \cite{ding2024joint}, ScanNet \cite{ding2024joint}, and MamabaSCD \cite{chen2024changemamba}. This figure clearly demonstrates that Semantic-CD generates more accurate segmentation masks, a reduction in semantic classification errors, and highlights the significant impact of semantic priors introduced by the CLIP model.

\begin{table}[!htbp]
  \centering
  \caption{Comparison with other state-of-the-art methods on the SECOND dataset.}
   \resizebox{0.85\linewidth}{!}{
    \begin{tabular}{l|rrrr}
    \toprule
    Methods & OA    & F1    & mIoU  & SeK \\
    \midrule
    HRSCD \cite{daudt2019multitask} & 83.88 & 47.69 & 70.58 & 16.81 \\
    ChangeMask  \cite{zheng2022changemask}& 85.53 & 50.54 & 71.39 & 18.36 \\
    SSCD \cite{ding2022bi}& 85.62 & 52.80  & 72.14 & 20.15 \\
    BiSRNet \cite{ding2022bi}& 85.80  & 53.59 & 72.69 & 21.18 \\
    TED \cite{ding2024joint}  & 85.50  & 52.61 & 72.19 & 20.29 \\
    SMNet \cite{niu2023smnet}& 84.29 & 50.48 & 71.62 & 18.98 \\
    ScanNet \cite{ding2024joint}& 86.13 & 55.21 & 72.48 & 21.57 \\
    MamabaSCD \cite{chen2024changemamba}& 90.36 & 55.32 & 73.68 & 22.92 \\
    \midrule
    baseline & 89.38 & 54.91 & 72.70  & 21.64 \\
    Semantic-CD & \textbf{91.31} & \textbf{56.11} & \textbf{75.10} & \textbf{23.85} \\
    \bottomrule
    \end{tabular}
    }
  \label{tab:comp}
\end{table}

 \subsection{Discussion}

This paper investigates an open-vocabulary approach to semantic change detection, enhancing its representation capabilities using the semantic priors of the CLIP foundation model. Due to the insufficient number of categories currently available for training and validation in remote sensing semantic change detection, this paper degrades the setting into a semantic change detection model with open capabilities, allowing all categories to participate in training, and the model itself exhibits open-vocabulary semantic segmentation capabilities. Under these conditions, Semantic-CD has demonstrated notable success, with its metrics significantly outperforming other methods, thereby validating the effectiveness of applying vision-language foundation models' (VLFMs) semantic priors in semantic change detection. In the future, we hope the release of more open-source datasets featuring diverse and granular categories in semantic change detection.

\section{Conclusion}

This paper introduces a method for semantic change detection, termed Semantic-CD, which features open vocabulary capabilities. Semantic-CD enhances the model's cross-category generalization by utilizing the implicit knowledge embedded in large-scale vision-language models (VLFMs). It predicts BCD and SCD results through a fully decoupled multi-task learning framework. Additionally, the proposed decoupled training process facilitates the fully extraction of latent knowledge from VLFMs, while minimizing interference between different task branches. Experimental results on the SECOND dataset show that Semantic-CD generates more precise mask regions and results in fewer semantic classification errors. This demonstrates its ability to effectively apply VLFMs' semantic priors to the SCD task.

\ifCLASSOPTIONcaptionsoff
  \newpage
\fi

\bibliographystyle{IEEEtran}
\small{
\bibliography{IEEEabrv,myreferences}
}

\end{document}